\documentclass{article}
\usepackage{spconf,amsmath, amsfonts,graphicx,hyperref, svg, booktabs, multirow, multicol}
\usepackage{comment, balance}%, ulem}
\title{Secure and reversible face anonymization with diffusion models}
\name{Pol Labarbarie$^1$, Vincent Itier$^2$ and William Puech$^1$%}
}
\address{$^1$ LIRMM, Univ Montpellier, CNRS, Montpellier, FRANCE\\ 
$^2$ IMT Nord Europe, Institut Mines-Télécom, Univ. Lille, Centre for Digital Systems, Lille, FRANCE}

\usepackage[dvipsnames]{xcolor}

\begin{document}
%\ninept
%
\maketitle
\begin{abstract}

Face anonymization aims to protect sensitive identity information by altering faces while preserving visual realism and utility for downstream computer vision tasks. Current methods struggle to simultaneously ensure high image quality, strong security guarantees, and controlled reversibility for authorized identity recovery at a later time. To improve the image quality of generated anonymized faces, recent methods have adopted diffusion models. However, these new diffusion-based anonymization methods do not provide a mechanism to restrict de-anonymization to trusted parties, limiting their real-world applicability. In this paper, we present the first diffusion-based framework for secure, reversible face anonymization via secret-key conditioning.
Our method injects the secret key directly into the diffusion process, enabling anonymization and authorized face reconstruction while preventing unauthorized de-anonymization. The use of deterministic forward and reverse diffusion steps guarantees exact identity recovery when the correct secret key is available. Experiments on CelebA-HQ and LFW demonstrate that our approach achieves better anonymization and de-anonymization capabilities than prior work. We also show that our method remains robust to incorrect or adversarial key de-anonymization. Our code will be made publicly available.

\end{abstract}

%
%%%%%%%%%%%%%%%%%%%%%%%%%%%%%%%%%%%%%%%%%%%%%%%%%%%%%%%%%%%%%%%%%%%%%%%%%%%%%
%                                   FUSION                                        %
%%%%%%%%%%%%%%%%%%%%%%%%%%%%%%%%%%%%%%%%%%%%%%%%%%%%%%%%%%%%%%%%%%%%%%%%%%%%%
\section{Introduction}
\label{sec:intro}

Computer vision systems are increasingly deployed in everyday applications, many of which involve the capture and processing of sensitive facial data. Face images reveal not only identity but also personal attributes such as age, gender, and emotional state, raising serious privacy concerns. Early anonymization techniques, such as Gaussian blurring or block-wise encryption~\cite{cichowski2011reversible}, aimed to obscure identity while retaining more or less useful visual information. Gaussian blurring preserves some degree of visual utility but provides only weak protection, whereas cryptographic transformations offer strong protection at the cost of rendering the protected image unusable and uninterpretable. These limitations have motivated the development of more advanced methods that seek to balance privacy, image quality, and utility for downstream vision tasks.

With the advent of deep learning, face anonymization methods have significantly advanced thanks to the improved generative capabilities of generative adversarial networks (GANs)~\cite{hukkelaas2019deepprivacy, maximov2020ciagan}. To reconstruct the original face only from anonymized faces and thus enable re-identification, several approaches develop a reverse process~\cite{gu2020password, li2023riddle, Yang2024G2face}. This reverse process, called de-anonymization, is essential for real-world applications such as criminal investigations using CCTV footage or testimonial videos. To guarantee that only authorized parties can reverse the anonymization, a secret key must constrain de-anonymization.  Secret-key conditioned reversible face anonymization was first introduced by Gu~\textit{et al.}~\cite{gu2020password}. In their approach, the input image (original or anonymized face) is concatenated with a pixel-replicated password (a binary image) and fed to an autoencoder-based network. To improve the visual quality of generated faces, Li~{\it et al.}~\cite{li2023riddle} introduce RiDDLE, a transformer-based Latent Encryptor module based on the StyleGAN2 model~\cite{karras2020analyzing}. The Latent Encryptor operates in the StyleGAN2 latent space, jointly processing the face latent code and a randomly sampled password at multiple scales using attention mechanisms. The resulting representations are fused and mapped to a single latent code, which is then fed to the StyleGAN2 generator to obtain the anonymized or de-anonymized face. To achieve more faithful de-anonymization, the anonymized latent code is saved as additional data during anonymization, raising concerns about vulnerability to attacks. To further improve the visual quality of the generated anonymized face, Yang~{\it et al.}~\cite{Yang2024G2face} add three additional modules to the StyleGAN2 model (G2Face method). Using data hiding techniques, the original face is embedded in the anonymized face for future face de-anonymization. However, no secret key is required for de-anonymization, leaving malicious actors free access. Yuan~\textit{et al.}~\cite{yuan2025ifadit} propose iFADIT, a dual-encoder framework that disentangles identity and attribute features. Identity features extracted using ArcFace~\cite{deng2019arcface} are transformed via a password-conditioned normalizing flow to produce anonymized identity representations. These are then combined with attribute features obtained from a pSp encoder~\cite{richardson2021encoding} and synthesized into anonymized faces using a StyleGAN generator. The use of an invertible flow enables controlled and reversible de-anonymization.
While GAN-based methods incorporate secret-key-controlled reversible anonymization, they remain limited by inherent GAN drawbacks, notably reduced generation diversity caused by mode collapse.

Diffusion models (DM)~\cite{ho2020denoising, song2020denoising, rombach2022high} have demonstrated remarkable advances in generative modeling, surpassing GAN in terms of image quality and generation diversity. Recent work have applied DM to perform face anonymization. Shaheryar {\it et al.}~\cite{shaheryar2024iddiffuse} develop a dual-conditional DM that, using a reference synthetic face as a conditioning, drives the anonymized face toward the synthetic face while preserving identity-irrelevant image features. Similarly, You {\it et al.}~\cite{you2024generation} use a latent DM conditioned by two additional face embeddings, one identity and one style embedding. These embeddings are used to anonymize the original face. Kung {\it et al.}~\cite{kung2025nullface} method includes also a face embeddings guidance and a mask guidance to anonymize face regions selectively. While these approaches can produce plausible anonymized faces, their anonymization procedures are either not reversible~\cite{shaheryar2024iddiffuse, kung2025nullface} or reversible only if the original face embeddings are retained~\cite{you2024generation}, which poses a significant security concern and requires additional data. Therefore, these methods are either not reversible or do not incorporate secret key conditioning, leaving them vulnerable to unauthorized de-anonymization. This absence of cryptographic constraints represents a significant gap in current research.

In this paper, we propose, to the best of our knowledge, the first diffusion-based reversible anonymization framework that integrates secret-key conditioning. Our approach builds upon a pre-trained unconditional DM, requiring neither model retraining nor model-specific modifications. The secret key is used to derive a sign-flipping vector that pseudo-randomly flips the components of the Gaussian noise realization for each face. To ensure a reversible one-to-one mapping between original and anonymized images, we rely on the deterministic DDIM process~\cite{song2020denoising}. Experiments conducted on CelebA-HQ~\cite{karras2017progressive} and LFW~\cite{huang2008labeled} datasets demonstrate that our method achieves both strong anonymization performance and accurate identity recovery through de-anonymization. We evaluate robustness against unauthorized de-anonymization attempts and show that successful identity recovery is possible only with the correct secret key, while even a single-bit deviation results in complete de-anonymization failure.

\begin{figure*}[hbtp!]
\centering
\includegraphics[width=\linewidth]{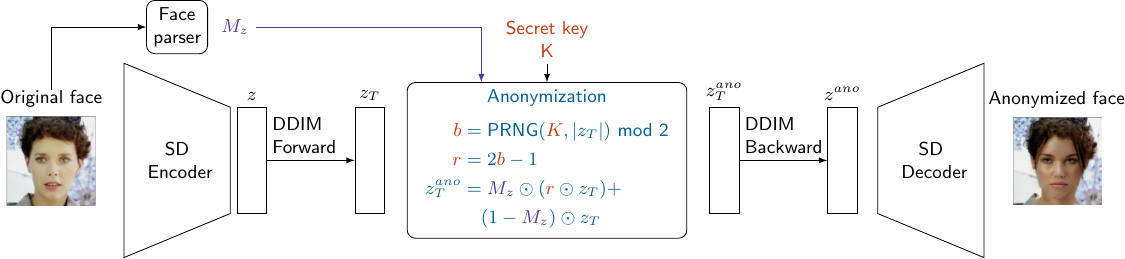}
%\includegraphics[width=0.70\linewidth]{figures/schema_end.png}
% \caption{Diagram of our method for face anonymization using a diffusion model. On top in blue and on bottom in green are represented the anonymization and de-anonymization procedure, respectively. SD is the abbreviation for Stable Diffusion.}
\caption{Diagram of our face anonymization pipeline. The original face is encoded by the Stable Diffusion (SD) encoder into a latent representation $z$ which is gradually noised over $T$ timesteps, using the DDIM deterministic forward process, to obtain its associated Gaussian realization $z_T$. A secret key $K$ is used to derive a Rademacher vector $r$ that anonymizes $z_T$, yielding $z_T^{ano}$. A rescaled facial mask $M_z$ is used to preserve identity irrelevant features. The anonymized Gaussian realization is deterministically denoised over $T$ timesteps to produce $z^{ano}$, which is decoded to generate the anonymized face $x^{ano}$.}
\label{method_scheme}
\end{figure*}

%%%%%%%%%%%%%%%%%%%%%%%%%%%%%%%%%%%%%%%%%%%%%%%%%%%%%%%%%%%%%%%%%%%%%%%%%%%%%
%                                                                           %
%%%%%%%%%%%%%%%%%%%%%%%%%%%%%%%%%%%%%%%%%%%%%%%%%%%%%%%%%%%%%%%%%%%%%%%%%%%%%
\section{Reversible anonymization with DM}
\label{sec:method}
%\vspace{-0.3cm}
Let $x_0 \in \mathbb{R}^{C\times H \times W}$ denote the original image, where $C$, $H$ and $W$ are the image channels, height and width respectively. In the case of the Denoising Diffusion Probabilistic Model (DDPM)~\cite{ho2020denoising}, the noising process, often denoted as forward diffusion process, can be expressed by:
\begin{equation}
    x_t = \sqrt{\frac{\alpha_t}{\alpha_{t-1}}}x_{t-1} + \sqrt{1 - \frac{\alpha_t}{\alpha_{t-1}}}\varepsilon_t , \quad \varepsilon_t \sim \mathcal{N}(0,I_d),
\label{equ:xt}
\end{equation}
where $x_t$ is the noised image and $\alpha_{1:T}\in  (0,1]^T$ is a decreasing sequence. When $t \to T$, $\alpha_t$ becomes sufficiently close to 0, we can show that the probability law of $x_T$ given $x_0$ converges to a standard Gaussian distribution~\cite{song2020denoising}. It is then natural to sample from a standard Gaussian distribution and run the backward process to obtain new images (see~\cite{ho2020denoising, song2020denoising} for backward process equations). A unified formulation of DM has been developed by Song~\textit{et al.}~\cite{song2020score} and a detailed introduction to DM is presented in~\cite{holderrieth2025introduction}. 

\subsection{Reversible anonymization through the Gaussian property}

\label{method_anonymization}
Our method takes advantage of the fact that $x_T$ is a realization of a standard Gaussian distribution in the following manner. 
For a Gaussian random variable $\varepsilon \sim \mathcal{N}(0, I_d)$, let $r \in \{-1,+1\}^d$ be a vector of independent variables that is independent of $\varepsilon$. Then, the element-wise product $r \odot \varepsilon$ is also a standard Gaussian distribution. Applied to our case, flipping any subset of dimensions of $x_T$ yields another valid Gaussian realization to generate a new valid image. We exploit this property to perform the proposed reversible face anonymization.

From a secret key $K$ and a pseudo-random number generator (PRNG), we generate a pseudo-random binary sequence $b \in \{0,1\}^d$, where $d$ is the dimension in which the diffusion is performed. This binary sequence is converted to a Rademacher vector (random vector containing -1 or 1), $r = 2b - 1 \in \{-1,+1\}^d $, which is then used to pseudo-randomly flip the coordinates of $x_T$, $x_T^{ano} = x_T \odot r$.

This anonymization procedure has the benefit of being fully reversible given the same secret key $K$ (without discrepancy), $x_T^{rec} = x_T^{ano} \odot r = x_T \odot ( r \odot r ) = x_T$, and ensure that $x_T$ and $x_T^{ano}$ are statistically indistinguishable from another standard Gaussian sample. Performing the de-anonymization with a wrong secret key $K'$ would result in a different Gaussian realization, $x_T^{wrong} =  x_T^{ano} \odot r' \neq  x_T $ that would generate a different, yet realistic, face image. The generation of a realistic, wrong de-anonymized face image is ensured by the fact that $x_T^{wrong}$ is still a realization of a standard Gaussian distribution, even though the secret key $K'$ is not the correct one.

\subsection{Deterministic forward and backward processes}

In order to use this Gaussian property for anonymization and de-anonymization presented in Section~\ref{method_anonymization}, we need to ensure that we obtain the same Gaussian realization $x_T$ given an image $x_0$ and that we obtain the same image $x'$ given a Gaussian realization $x_T'$. In other words, we need the forward and backward diffusion processes to be both deterministic and reversible. The Denoising Diffusion Implicit Model (DDIM) method~\cite{song2020denoising} is a standard diffusion sampling method that allows reversibility. When the DDIM stochastic parameter $\sigma_t = 0$ (see Equ.~12 in~\cite{song2020denoising}), the DDIM backward process becomes deterministic and is given by:
\begin{equation}\label{ddim_backward_deterministic}
\resizebox{0.91\columnwidth}{!}{$
x_{t-1} =
\sqrt{\frac{\alpha_{t-1}}{\alpha_t}} \, x_t
+ \left(
\sqrt{\frac{1}{\alpha_{t-1}} - 1} - \sqrt{\frac{1}{\alpha_t} - 1}
\right)
\cdot \varepsilon_\theta(x_t, t),
$}
\end{equation}
where $\varepsilon_\theta(x_t, t)$ is the neural network (often called a denoiser) already trained that, given a noisy face $x_t$ at time $t$, estimates the noise level. Based on the assumption that the ordinary differential equation process can be reversed in the limit of small steps~\cite{song2020denoising} a forward process can be constructed as the inverse of the deterministic DDIM backward process. This process, called the DDIM forward process, is:
\begin{equation}\label{ddim_forward}
\resizebox{0.91\columnwidth}{!}{$
x_{t+1} = 
\sqrt{\frac{\alpha_{t+1}}{\alpha_t}} \, x_t
+ \left( \sqrt{\frac{1}{\alpha_{t+1}} - 1} - \sqrt{\frac{1}{\alpha_t} - 1} \right) 
\cdot \varepsilon_\theta(x_t, t).
$}
\end{equation}
Taken together, this yields a fully deterministic pipeline. We use Eq.~\eqref{ddim_forward} to map a face image $x_0$ to a Gaussian realization $x_T$, then we use Eq.~\eqref{ddim_backward_deterministic} to reconstruct $x_0$ from $x_T$ in a deterministic manner.

\subsection{Face anonymization overall pipeline}

An overview of the proposed method is shown in Fig.~\ref{method_scheme}. We adopt the widely used Stable Diffusion (SD) model~\cite{rombach2022high}, where the forward diffusion process operates on latent encoding $z_0 = E(x_0)$, and the final reconstruction is obtained through the image decoder $ x_0 = D(z_0)$ at the end of the backward process, where $E(.)$ and $D(.)$ are the encoder and decoder respectively. Despite being denoted as a latent vector, $z_t$ is usually a tensor of shape $c \times h \times w$, where $c$ is the channel dimension and $h \times w$ are the spatial dimensions. Any element-wise product between a vector of size $h \times w$ and a latent vector $z_t$ is performed by broadcasting the vector over the corresponding dimensions of $z_t$.

To ensure that our method preserves the identity-irrelevant image features, we extract the facial mask of the face using a face parser. This facial mask is rescaled to match the spatial dimensions of SD latent space $h \times w$ and is denoted $M_z$. We use SD encoder followed by Eq.~\eqref{ddim_forward} to map the original face $x_0$ to its associated Gaussian realization $z_T$. Then, our anonymization procedure described in Section~\ref{method_anonymization} is applied to the masked regions, leaving the remaining elements unchanged. Formally, the anonymized latent representation is given by:
\begin{equation}
z_T^{ano} = M_z \odot (r \odot z_T) + (1-M_z) \odot z_T.
\end{equation}
Then we run the DDIM backward process (Eq. \ref{ddim_backward_deterministic}) and at each timestep we re-inject the identity-irrelevant image feature:
\begin{equation}
z_t^{ano} = M_z \odot z_t^{ano} + (1-M_z) \odot z_t,
\end{equation}
where $z_t$ was obtained previously during the DDIM forward process of the original face latent. These intermediate latents, $z_t$, are erased once the anonymization procedure is complete. Finally, using the SD decoder, we obtain the anonymized face $x^{ano}$ (see the top in blue of Fig.~\ref{method_scheme}).

\subsection{De-anonymization procedure}

For the de-anonymization, only the anonymized face image $x^{ano}$ and the secret key $K$ are necessary. The de-anonymization procedure follows the anonymization procedure. The rescaled facial mask, $M_{z^{ano}}$, is extracted from the anonymized face $x^{ano}$ using a face parser. Using SD encoder and Eq.~\eqref{ddim_forward}, the anonymized face image is mapped to its associated Gaussian realization $z_T^{ano}$. The de-anonymization procedure described in Section~\ref{method_anonymization} is applied to the masked regions, while the remaining elements are also left unchanged:
\begin{equation}
z_T^{rec} = M_{z^{ano}} \odot (r \odot z_T^{ano}) + (1-M_{z^{ano}}) \odot z_T^{ano}.
\end{equation}
The DDIM backward process is run (Eq.~\eqref{ddim_backward_deterministic}) and at each timestep the identity-irrelevant image features from the anonymized face are re-injected:
\begin{equation}
z_t^{rec} = M_{z^{ano}} \odot z_t^{rec} + (1-M_{z^{ano}}) \odot z_t^{ano},
\end{equation}
where $z_t^{ano}$ was obtained previously during the DDIM forward process of the anonymized face latent.  The de-anonymized face $x^{rec}$ is reconstructed using SD decoder~$D$.

%%%%%%%%%%%%%%%%%%%%%%%%%%%%%%%%%%%%%%%%%%%%%%%%%%%%%%%%%%%%%%%%%%%%%%%%%%%%%
%                                                                           %
%%%%%%%%%%%%%%%%%%%%%%%%%%%%%%%%%%%%%%%%%%%%%%%%%%%%%%%%%%%%%%%%%%%%%%%%%%%%%
\vspace*{-0.5cm}
\section{Experiments \& Results}
\label{sec:results}

\subsection{Experimental settings}

\textbf{Evaluated methods.} We compare our method with the open-sourced reversible face anonymization approaches that are constrained by a secret key, namely RiDDLE~\cite{li2023riddle}, G2-Face~\cite{Yang2024G2face} and iFADIT ~\cite{yuan2025ifadit}. RiDDLE (A.D.) stands for RiDDLE with additional data. We use publicly shared model weights and code for these methods.

\noindent  \textbf{Datasets.} We follow the experimental settings of previous work~\cite{gu2020password, li2023riddle, Yang2024G2face, yuan2025ifadit}. The method components are trained, if necessary, on the FFHQ dataset~\cite{karras2019style}, which comprises 70,000 face images. Evaluations are performed on the LFW~\cite{huang2008labeled} and CelebA-HQ datasets~\cite{karras2017progressive}, which consist of 1600 pairs of images and 30,000 images, respectively. In the LFW dataset, pairs of images are different images of the same identity, whereas in CelebA-HQ, all images are from different identities. Following~\cite{Yang2024G2face, yuan2025ifadit}, all the images are resized and cropped to the size of $256 \times 256$.

\noindent \textbf{Evaluation metrics.} To evaluate the degree of anonymization and de-anonymization achieved by the different methods, we measure either cosine similarity or the true acceptance rate (TAR), depending on the dataset. For the CelebA-HQ dataset, we report cosine similarity scores, as a reference template set cannot be constructed. For the LFW dataset, we apply the anonymization methods to the second image of each image pair and use the corresponding first image to build a template gallery for face identification. Before anonymization, a cosine similarity threshold is derived from the second images such that the false acceptance rate (FAR) does not exceed $0.1\%$. We evaluate the true acceptance rate for this designed threshold, often denoted TAR@FAR=0.1\%. For all evaluations, we report results using three face recognition models: FaceNet~\cite{schroff2015facenet} trained on CASIA, ArcFace~\cite{deng2019arcface} trained on MS1MV3, and AdaFace~\cite{kim2022adaface} trained on WebFace12M.

\noindent \textbf{Our method settings.} We adopt the unconditional DM from the SD paper~\cite{rombach2022high}. This model has already been trained on the FFHQ dataset. Model weights are available \href{https://github.com/CompVis/latent-diffusion?tab=readme-ov-file#model-zoo}{here}. For anonymization and reconstruction, we use the deterministic DDIM forward and reverse processes with 
$T=50$ timesteps, following common practice in the diffusion modeling literature~\cite{song2020denoising, song2020score}. This choice provides a trade-off between computational efficiency and image quality. Facial masks are obtained using the BiSeNet face parser~\cite{yu2018bisenet}. We sample a 128-bit secret key $K$, \textit{i.e.} $K \in \{0,...,2^{128}\}$, and this secret key is used to generate $r \in \{-1,+1\}^{d}$ where $d = 3 \times 64 \times 64$ for the model we use.

\begin{table}[t]
\centering
%\scalebox{0.86}{
\resizebox{0.99\columnwidth}{!}{%
\begin{tabular}{lccc}
\toprule
\multirow[c]{2}{*}{Method}   & \textbf{FaceNet} & \textbf{ArcFace-R50} & \textbf{AdaFace-ViT} \\
  & \textbf{(CASIA)} & \textbf{(MS1MV3)} & \textbf{(WebFace12M)} \\
\hline
\multicolumn{4}{c}{\textbf{Anonymization} (lower is better ↓)}\\
\hline
RiDDLE~\cite{li2023riddle} &
 \textbf{0.092 $\pm$ 0.150} & 0.042 $\pm$ 0.080 & 0.074 $\pm$ 0.080 \\

G2Face~\cite{Yang2024G2face}  &
 0.157 $\pm$ 0.146 & \underline{0.037 $\pm$ 0.096} & \underline{0.061 $\pm$ 0.094} \\

iFADIT~\cite{yuan2025ifadit} &
 0.290 $\pm$ 0.153 & 0.120 $\pm$ 0.093 & 0.159 $\pm$ 0.091 \\

\textbf{Ours} &
 \underline{0.093 $\pm$ 0.140} & \textbf{0.029 $\pm$ 0.077} & \textbf{0.049 $\pm$ 0.078} \\
\hline\hline
\multicolumn{4}{c}{\textbf{De-Anonymization} (higher is better ↑)}\\
\hline
RiDDLE~\cite{li2023riddle} &
 0.181 $\pm$ 0.151 & 0.076 $\pm$ 0.084 & 0.11 $\pm$ 0.083 \\

RiDDLE (A.D.)~\cite{li2023riddle} &
 \underline{0.745 $\pm$ 0.083} & 0.528 $\pm$ 0.098 & 0.538 $\pm$ 0.086 \\

G2Face~\cite{Yang2024G2face}  &
\textbf{0.783 $\pm$ 0.078} & \underline{0.680 $\pm$ 0.090} & \underline{0.671 $\pm$ 0.077} \\

iFADIT~\cite{yuan2025ifadit} &
 0.290 $\pm$ 0.141 & 0.119 $\pm$ 0.096 & 0.159 $\pm$ 0.090 \\

\textbf{Ours} &
 0.743 $\pm$ 0.182 & \textbf{0.692 $\pm$ 0.201} & \textbf{0.707 $\pm$ 0.180} \\
\hline
\end{tabular}}
\caption{Cosine similarities for anonymization (between original and anonymized faces) and for de-anonymization (between original and de-anonymized faces). Cosine similarities are evaluated for several face recognition models on the CelebA-HQ dataset. Bold and underlined values indicate the best and second-best results, respectively.}
\label{tab:cosine_similarity}
\end{table}

\begin{figure}[t]
\centering
\begin{tabular}{ccccc}
    \multicolumn{5}{c}{\includegraphics[width=0.94\linewidth]{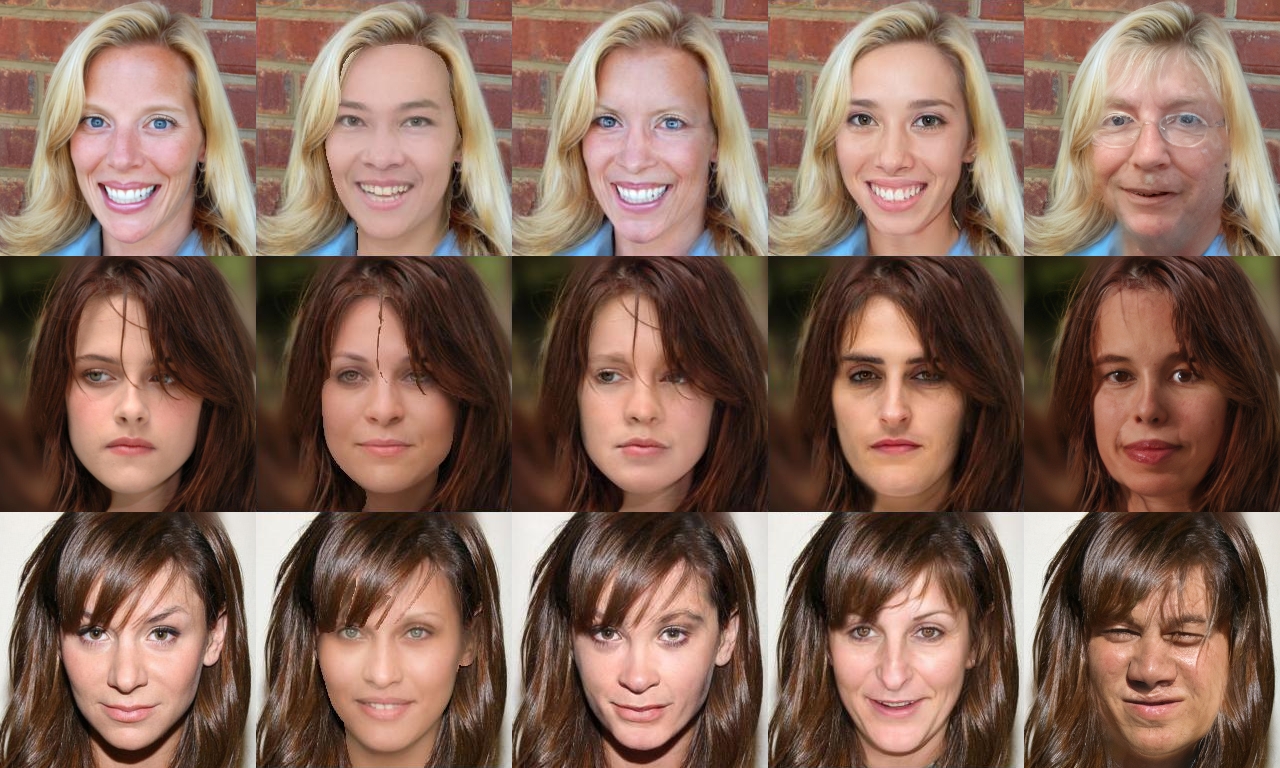}} \\
    Original  &  RiDDLE  & G2 & iFADIT & \multirow[c]{2}{*}{Ours} \\
        face &  (A.D.)~\cite{li2023riddle} & Face~\cite{Yang2024G2face}  & ~\cite{yuan2025ifadit} & \\
\end{tabular}
    \caption{Qualitative comparison of face anonymization among different methods on the CelebA-HQ dataset.}
    \vspace{-0.5cm}
    \label{fig:anonymization}
\end{figure}

\begin{table}[t]
\centering
%\scalebox{0.86}{
\resizebox{0.99\columnwidth}{!}{%
\begin{tabular}{lccc}
\toprule
\multirow[c]{2}{*}{Method}  & \textbf{FaceNet} & \textbf{ArcFace-R50} & \textbf{AdaFace-ViT} \\
 & \textbf{(CASIA)}  & \textbf{(MS1MV3)} & \textbf{(WebFace12M)} \\
\hline
Clean & 90.96\% & 99.06\% & 99.62 \\
\hline 
\hline
\multicolumn{4}{c}{\textbf{Anonymization} (lower is better ↓)}\\
\hline
RiDDLE~\cite{li2023riddle} &
0.314\% & \underline{0.628\%} & \textbf{0.384\%} \\

G2Face~\cite{Yang2024G2face} &
 \textbf{0.105\%} & 0.628\% & 1.082\% \\

iFADIT~\cite{yuan2025ifadit} &
 1.152\% & 1.711\% & 1.187\% \\

\textbf{Ours} &
 \underline{0.279\%} & \textbf{0.559\%} & \underline{0.489\%} \\
\hline\hline
\multicolumn{4}{c}{\textbf{De-Anonymization} (higher is better ↑)}\\
\hline
RiDDLE~\cite{li2023riddle} &
 0.279\% & 0.489\% & 0.803\% \\

RiDDLE (A.D.)~\cite{li2023riddle} &
 20.985\% & 29.853\% & 33.904\% \\

G2Face~\cite{Yang2024G2face} &
 \underline{24.825\%} & \textbf{76.257\%} & \textbf{79.888\%} \\

iFADIT~\cite{yuan2025ifadit} &
 0.314\% & 1.047\% & 0.943\% \\

\textbf{Ours} &
 \textbf{38.757\%} & \underline{70.531\%} & \underline{70.216\%} \\

\hline
\end{tabular}
}
\caption{TAR@FAR=0.1\% for anonymization and de-anonymization. TAR are evaluated for several face recognition models on the LFW dataset.}
\label{tab:tar}
\end{table}

%%%%%%%%%%%%%%%%%%%%%%%%%%%%%%%%%%%%%%%%%%%%%%%%%%%%%%%%%%%%%%%%%%%%%%%%%%%%%
\subsection{Face anonymization}
\label{sec:obs-rec}
 The first part of Table~\ref{tab:cosine_similarity} and~\ref{tab:tar} summarize the anonymization results. Our method achieves the lowest or second-lowest cosine similarity score and TAR@FAR=0.1\% among all compared approaches. These low scores indicate a superior level of identity obscuration, confirming that the anonymized faces are significantly different from the originals. Qualitative face anonymization results are shown in Fig.~\ref{fig:anonymization}. Our method generates high-quality, realistic anonymized faces while preserving identity-irrelevant attributes such as background, hair, and pose. The visual quality of the generated faces is comparable to or exceeds that of the previous methods, demonstrating the strong generative capabilities of the diffusion model. Our method seems to produce a greater diversity among the anonymized faces.

\subsection{De-anonymization: original face recovery}
\label{sec:obs-rec}

\begin{figure}[t]
\centering
\begin{tabular}{ccccc}
    \multicolumn{5}{c}{\includegraphics[width=1\linewidth]{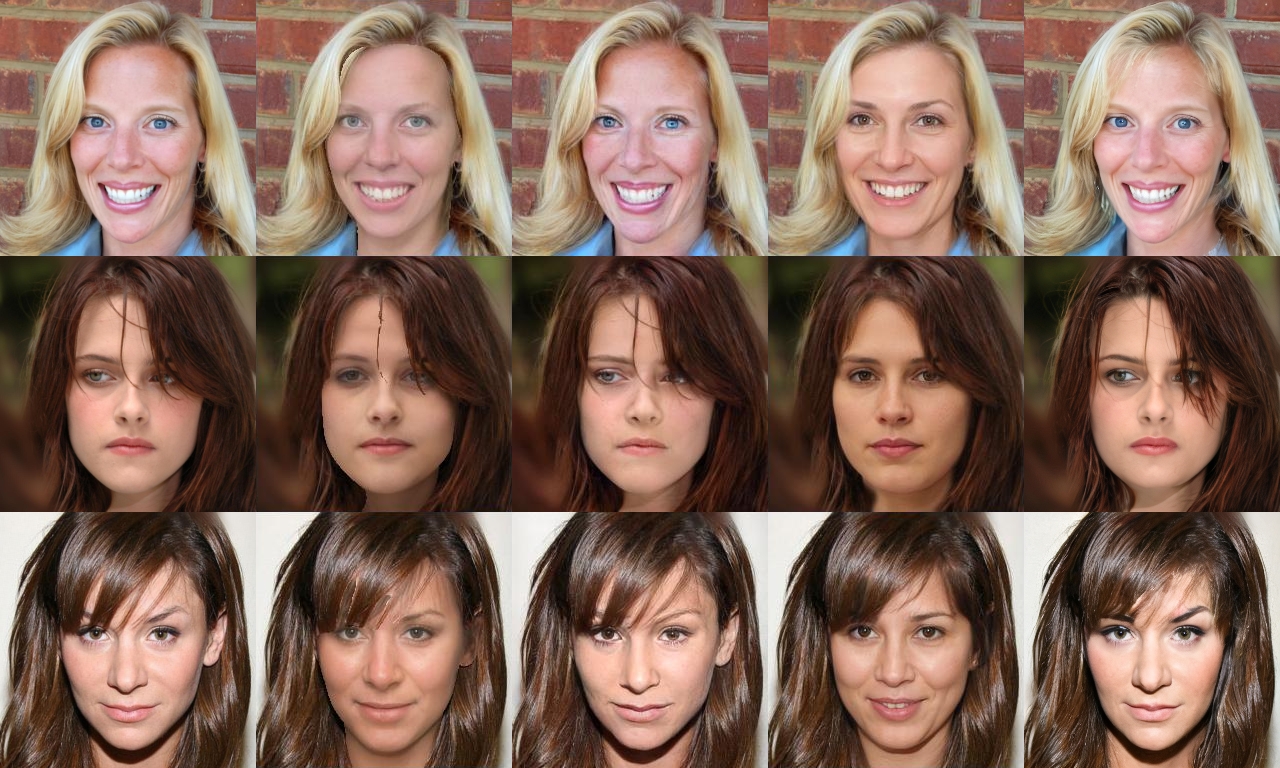}} \\
    Original  &  RiDDLE  & G2 & iFADIT & \multirow[c]{2}{*}{Ours} \\
        face &  (A.D.)~\cite{li2023riddle} & Face~\cite{Yang2024G2face}  & ~\cite{yuan2025ifadit} & \\
\end{tabular}
    \caption{Qualitative comparison of face de-anonymization among different methods on the CelebA-HQ dataset.}
    \label{fig:deanonymization}
    \vspace{-0.3cm}
\end{figure}

Face de-anonymization results are provided in the second part of Table~\ref{tab:cosine_similarity} and~\ref{tab:tar}. As for the face anonymization results, our method gets the best or second-best cosine similarity score and TAR@FAR=0.1\%, demonstrating its ability to recover the original face and thus re-identify people. Fig.~\ref{fig:deanonymization} displays some qualitative results of the de-anonymization process. The recovered faces generated by our method demonstrate high fidelity to the original faces, allowing re-identification. As highlighted by both numerical and qualitative results, the iFADIT method~\cite{yuan2025ifadit} fails to reconstruct a face that enables the re-identification of the original face. RiDDLE~\cite{li2023riddle} has to rely on additional data to reconstruct a face enabling re-identification, increasing method complexity and security leaks.

\subsection{Wrong password de-anonymization}

%%% REPLACE 
%$K_{true}$ -> $K_{t}$
%$K_{t} \oplus \hat K_{1bit}$ -> $\hat K_{1}$
%$K_{random}$ -> $K_{r}$

\begin{figure}
    \centering
    \includegraphics[width=1\linewidth]{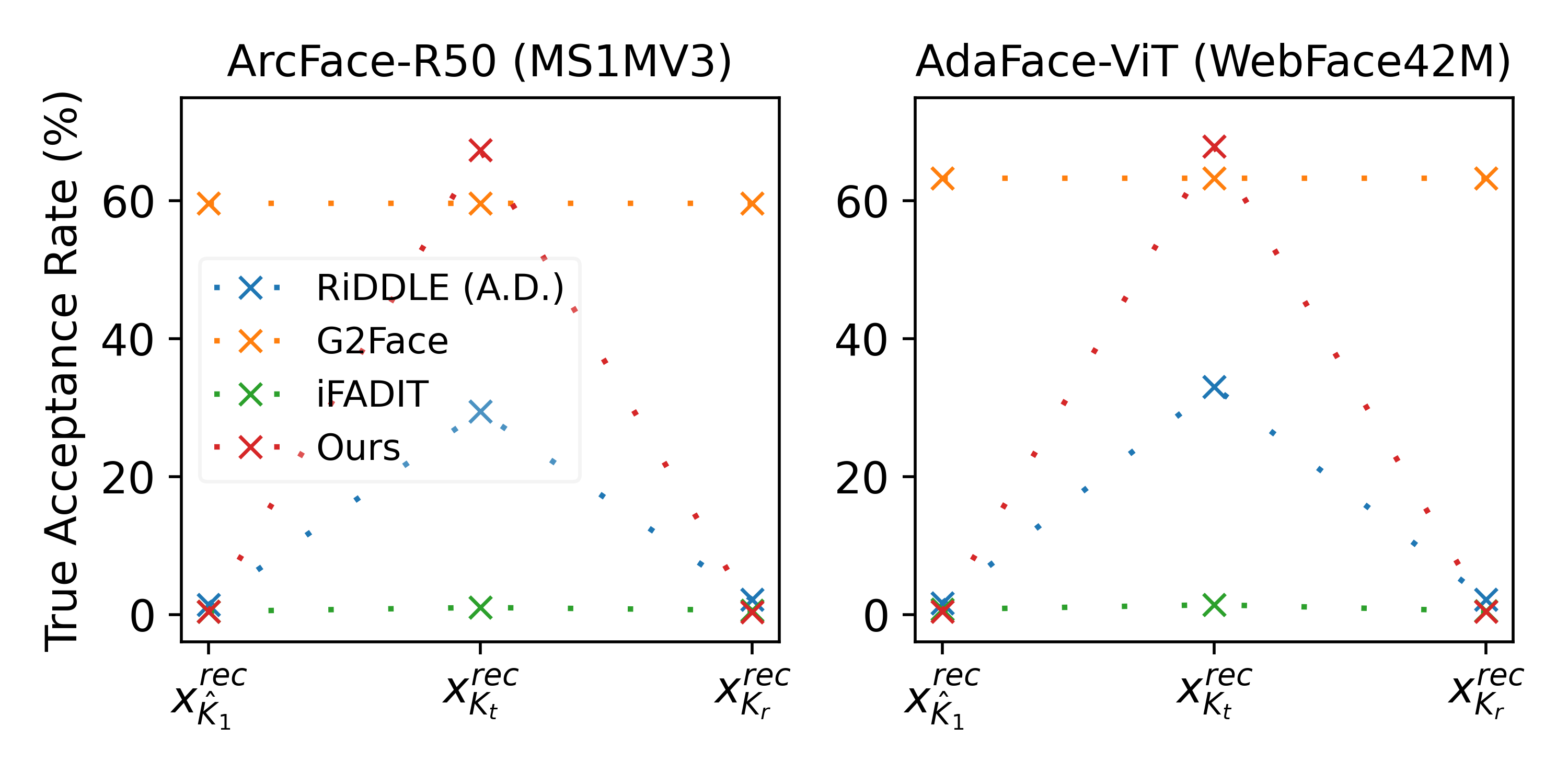}
        \vspace{-0.64cm}
    \caption{TAR@FAR=0.1\% under different de-anonymization scenarios. TAR are evaluated for several face recognition models on the LFW dataset.}
    \label{fig:tar:wrong:secret:key}
\end{figure}

\begin{figure}[t]
\includegraphics[width=1.\linewidth]{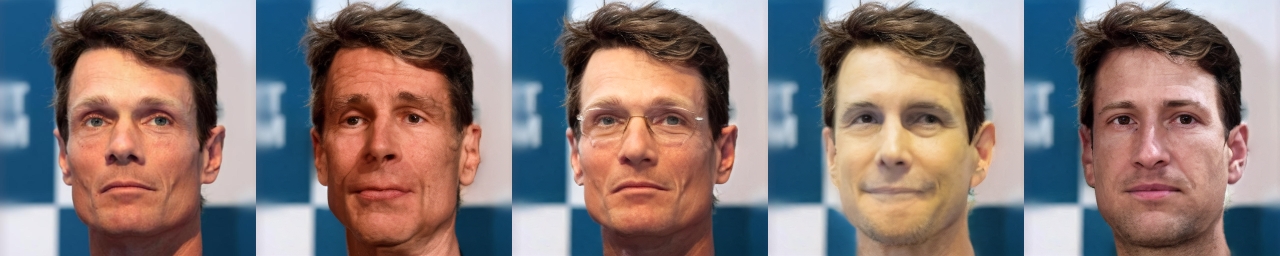}

\vspace{0.2cm}

\begin{minipage}{0.18\linewidth}
    \centering
    Original face
\end{minipage}\hfill
\begin{minipage}{0.18\linewidth}
    \centering
    $x^{ano}$
\end{minipage}\hfill
\begin{minipage}{0.18\linewidth}
    \centering
    $x^{rec}_{K_{t}}$
\end{minipage}\hfill
\begin{minipage}{0.18\linewidth}
    \centering
    $x^{rec}_{\hat{K}_{1}}$
\end{minipage}\hfill
\begin{minipage}{0.18\linewidth}
    \centering
    $x^{rec}_{K_{r}}$
\end{minipage}
\caption{Qualitative comparison of de-anonymized faces under different de-anonymization scenarios on the CelebA-HQ dataset.}
\label{fig:wrong_ano}
\end{figure}

In this section, we analyze the robustness of methods against unauthorized de-anonymization attempts. We consider three attack scenarios: (i) de-anonymization using the same secret-key as that employed for anonymization (denoted $K_{t}$), (ii) de-anonymization using a secret key that differs by a single bit from the correct key (denoted $\hat{K}_{1}$) and, (iii) de-anonymization using a randomly sampled secret key (denoted $K_{r}$). Fig.~\ref{fig:tar:wrong:secret:key} reports the TAR@FAR=0.1\% under these three scenarios. Providing a wrong secret key leads to a complete failure in recovering the original identity for all methods except G2Face~\cite{Yang2024G2face}, which remains vulnerable to malicious de-anonymization and potential identity leakage. Fig.~\ref{fig:tar:wrong:secret:key} shows that, when the correct secret key is provided, our method achieves superior de-anonymization performance compared to RiDDLE~\cite{li2023riddle}, despite the latter having access to additional data during de-anonymization. Fig.~\ref{fig:wrong_ano} illustrates that, under an incorrect secret-key, our method produces either a distinctly different anonymized identity or a severely corrupted image, thereby preventing an attacker from determining whether de-anonymization was successful.

% \begin{figure}[t]
% \centering
% \begin{tabular}{ccccc}
%     \multicolumn{5}{c}{\includegraphics[width=0.94\linewidth]{figures/icip/wrong_reco.jpg}} \\
%     Original  &  \multirow[c]{2}{*}{$x^{ano}$}  & \multirow[c]{2}{*}{$x^{rec}_{K_{t}}$} & \multirow[c]{2}{*}{$x^{rec}_{\hat{K}_{1}}$} & \multirow[c]{2}{*}{$x^{rec}_{K_{r}}$} \\
%         face &   &   &  & \\
% \end{tabular}
% \vspace{-0.2cm}
%     \caption{Qualitative comparison of de-anonymized faces under different de-anonymization scenarios on the CelebA-HQ dataset. \corV{fixer les légendes}}
%     \label{fig:wrong_ano}
% \end{figure}

% \begin{figure}[hbtp!]
% \centering
% \setlength{\tabcolsep}{4pt}
% \begin{tabular}{c|c|c|cc}
%     \includegraphics[width=0.16\linewidth]{figures/wrong2/ori.png} &
%     \includegraphics[width=0.16\linewidth]{figures/wrong2/ano.png} &
%      \includegraphics[width=0.16\linewidth]{figures/wrong2/reco.png} &
%     \includegraphics[width=0.16\linewidth]{figures/wrong2/wrong1.png} &
%     \includegraphics[width=0.16\linewidth]{figures/wrong2/wrong2.png} \\ 
%     Original face & $x^{ano}|k_1$ & $x^{rec}_{k_1}|k_1$ &$x^{rec}_{k_1}|k_2$ & $x^{rec}_{k_1}|k_3$ 
% \end{tabular}
%     \caption{Qualitative comparison of de-anonymized faces under different de-anonymization scenarios on the CelebA-HQ dataset.}
%     \label{fig:wrong_ano}
% \end{figure}

%%%%%%%%%%%%%%%%%%%%%%%%%%%%%%%%%%%%%%%%%%%%%%%%%%%%%%%%%%%%%%%%%%%%%%%%%%%%%
%                                                                           %
%%%%%%%%%%%%%%%%%%%%%%%%%%%%%%%%%%%%%%%%%%%%%%%%%%%%%%%%%%%%%%%%%%%%%%%%%%%%%
% \vspace{-0.2cm}
\section{Conclusion}
\label{sec:conclusion}
\vspace{-0.2cm}

In this paper, we propose the first key-conditioned reversible face anonymization framework based on a pre-trained diffusion model. Our approach manipulates the noisy latent space using a secret key and employs a deterministic DDIM process with facial masking to enable controlled and reversible anonymization. 
The anonymization process, based on a pseudo-randomized flip of the latent face representation, allows us to evolve in the space of representations learned by the model and thus produce high-quality, diverse faces. 
Our method achieves state-of-the-art anonymization and de-anonymization while producing wrong faces under incorrect secret keys. These different findings illustrate that our method achieves a good trade-off between security enforcement and de-anonymization performance, while producing high-quality face anonymization.

\clearpage
\balance
\section*{ACKNOWLEDGMENT}
\vspace{-0.24cm}

This work was supported by a French government funding grant managed by the Agence National de la Recherche under the France 2030 program, reference ANR-22-PECY-0011. This work was performed using HPC resources from GENCI-IDRIS (Grant 20AD-011016646)
\vspace{-0.3cm}

\bibliographystyle{IEEEtran}

\bibliography{bibliography}

\end{document}